\documentclass[10pt,twocolumn]{article} 
\usepackage{arxivTwoColumnStyle}
\usepackage[round]{natbib}
\renewcommand{\bibname}{References}

\newif\ifaistat

\aistatfalse

\newcommand{\aistatCond}[2]{\ifaistat #1 \else #2 \fi}

\usepackage{subfig}
\usepackage[shortlabels]{enumitem}
\usepackage[dvipsnames,table,xcdraw]{xcolor}

\usepackage{ifpdf}

\usepackage{amsmath,amsfonts,bm}

\def\eqref#1{equation~\ref{#1}}

\def\floor#1{\lfloor #1 \rfloor}
\def\1{\bm{1}}

\DeclareMathAlphabet{\mathsfit}{\encodingdefault}{\sfdefault}{m}{sl}
\SetMathAlphabet{\mathsfit}{bold}{\encodingdefault}{\sfdefault}{bx}{n}

\usepackage[utf8]{inputenc} 
\usepackage[T1]{fontenc}    

\usepackage{enumitem}  
\usepackage{xr}
\usepackage{hyperref}       
\usepackage{booktabs}       
\usepackage{amsfonts}       
\usepackage{nicefrac}       
\usepackage{microtype}      
\usepackage{lipsum}		
\usepackage{enumitem}
\usepackage{wrapfig}
\usepackage{bbold}
\usepackage{graphicx}

\usepackage{fancybox}
\usepackage{listings}
\usepackage{amssymb}
\usepackage{booktabs}
\usepackage[noend,algoruled,linesnumbered, rightnl]{algorithm2e}
\usepackage{seqsplit}
\usepackage{multirow}
\usepackage{float}
\usepackage{doi}
\usepackage{pgfplots}
\usepackage{pgf}
\usepackage{calc}  
\usepackage{array}

\usepackage{mathtools}
\usepackage{xcolor}
\usepackage{bbm}
\usepackage{multicol}
\usepackage{caption}
\usepackage[draft]{minted}
\input{_minted-arxiv_version/default-pyg-prefix.pygstyle}
\input{_minted-arxiv_version/default.pygstyle}

\usepackage{accents}

\renewcommand{\url}[1]{\texttt{\seqsplit{#1}}}

\usepackage{tikz}
\usepackage{adjustbox}

\makeatletter
\newcommand*{\addFileDependency}[1]{
  \typeout{(#1)}
  \@addtofilelist{#1}
  \IfFileExists{#1}{}{\typeout{No file #1.}}
}
\makeatother

\newcommand{\comment}[1]{}
\long\def\/*#1*/{}

{\begin{Sbox}\begin{minipage}}%
{\end{minipage}\end{Sbox}\fbox{\TheSbox}}

\DeclareSymbolFont{bbold}{U}{bbold}{m}{n}
\DeclareSymbolFontAlphabet{\mathbbold}{bbold}
\SetKwInput{KwInput}{Input} 
\SetKwInput{KwOutput}{Output} 
\definecolor{lightergray}{RGB}{242, 242, 242}

\newcommand\inputpgf[2]{{
\let\pgfimageWithoutPath\pgfimage
\renewcommand{\pgfimage}[2][]{\pgfimageWithoutPath[##1]{#1/##2}}
\let\includegraphicsWithoutPath\includegraphics
\renewcommand{\includegraphics}[2][]{\includegraphicsWithoutPath[##1]{#1/##2}}
\input{#1/#2}
}}

\newcommand{\quotes}[1]{``#1''}

\newcommand{\ourlib}{\href{https://github.com/mlzxy/qsparse}{https://github.com/mlzxy/qsparse}}

\newcolumntype{T}{>{\tiny}c}
\newcolumntype{S}{>{\scriptsize}c}

\newcommand{\script}[1]{{\scriptsize #1}}

\newcommand{\pq}[2]{{\small $P_{0.5}(#1)\rightarrow Q_8(#2)$}}

\newcommand{\qp}[2]{{\small $Q_8(#1)\rightarrow P_{0.5}(#2)$}}

\newcommand{\q}[1]{{\small $Q_8(#1)$}}

\newcommand*{\mat}[1]{\mathbf{#1}}

\newcommand{\ian}{\textcolor{blue}}

\newcommand{\xinyu}[1]{\textcolor{brown}{#1}}

\newcommand{\ianUpdate}{}
\newcommand{\icc}{}
\newcommand{\Question}{}

\renewcommand{\ian}{}
\renewcommand{\xinyu}{}

\newcommand{\ianLast}{}

\begin{document}

\ifaistat
	\twocolumn[

		\aistatstitle{Training Deep Neural Networks with Joint Quantization and Pruning of Weights and Activations}

		\aistatsauthor{ Xinyu Zhang* \And Ian Colbert* \And  Ken Kreutz-Delgado \And Srinjoy Das }

		\aistatsaddress{UC San Diego \And  UC San Diego \And UC San Diego \And  West Virginia University} 
	]
\else

    \title{Training Deep Neural Networks with Joint Quantization and Pruning of Weights and Activations}
    
    \author{
      Xinyu Zhang\textsuperscript{1,}\thanks{Equal contribution.}\\
      \texttt{xiz368@eng.ucsd.edu}
      \and
      Ian Colbert\textsuperscript{1,}\footnotemark[1]\\
      \texttt{icolbert@eng.ucsd.edu}
      \and
      Ken Kreutz-Delgado\textsuperscript{1}\\
      \texttt{kreutz@eng.ucsd.edu}
      \and
      Srinjoy Das\textsuperscript{2}\\
      \texttt{srinjoy.das@mail.wvu.edu} \\
    }

\fi


\hypersetup{
pdftitle={Training Deep Neural Networks with Joint Quantization and Pruning of Weights and Activations},
pdfsubject={deep learning},
pdfauthor={Xinyu Zhang},
pdfkeywords={Quantization, Neural Networks, Pruning, Feature Pruning, Software Tool, Python, Pytorch, 
Model Compression, Object Detection, Super Resolution, Classification, Image to Image
Translation, Generative Adversarial Network, CycleGAN},
}

\maketitle

\begin{abstract}

Quantization and pruning are core techniques used to reduce the inference costs of deep neural networks.
State-of-the-art quantization techniques are currently applied to both the weights and activations; however, pruning is most often applied to only the weights of the network.
{In this work, we jointly apply novel uniform quantization and unstructured pruning methods to both the weights and activations of deep neural networks during training.
Using our methods, we empirically evaluate the currently accepted prune-then-quantize paradigm across a wide range of computer vision tasks and observe a non-commutative \xinyu{relationship between pruning and quantization} when applied to both the weights and activations of deep neural networks.
Informed by these observations, we articulate the \textit{non-commutativity} hypothesis: for a given deep neural network being trained for a specific task, there exists an exact training schedule in which quantization and pruning can be introduced to optimize network performance.
We identify that this optimal ordering \ianUpdate{not only exists, but also} varies across discriminative and generative tasks\ianUpdate{.
Using the optimal \icc{training schedule} within our training framework, we} demonstrate increased performance per memory footprint over existing solutions.}
\end{abstract}
	
\begin{figure*}[t]
	\centering
	\includegraphics[width=\linewidth]{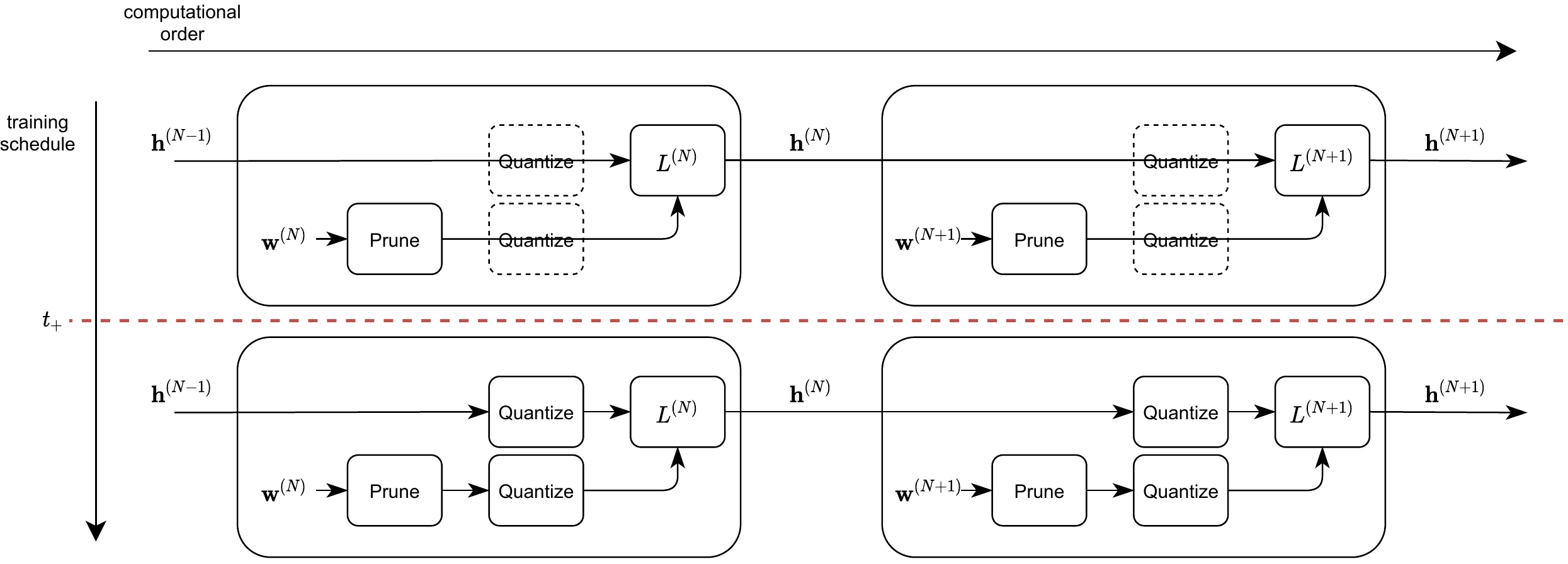}
    \caption{\small{\ianUpdate{\icc{The majority of} state-of-the-art frameworks currently default to a ``prune-then-quantize" \icc{training schedule} when jointly applying quantization and pruning to \icc{the weights of} DNNs~\citep{n8_han2015deep, n10_zhao2019focused, n17_yu2020joint}.
    \icc{In such a paradigm, the quantization operator is inactive in the computational order (horizontal flow) until time $t_+$, when it is introduced into by the training schedule (vertical flow) to limit precision.}
    \icc{Note that} in existing approaches, while quantization techniques are applied to both the weights ($\mathbf{w}^{(i)}$) and activations ($\mathbf{h}^{(i)}$) before executing the forward pass of the layer ($L^{(i)}$), pruning is most often only applied to the weights of the network.}}}
	\label{fig:prune_then_quantize}
\end{figure*}

\footnotetext[1]{Department of Electrical and Computer Engineering, University of California, San Diego.}
\footnotetext[2]{School of Mathematical and Data Sciences, West Virginia University.}
	
\section{Introduction}
\label{sec:intro}

\ian{The performance of deep neural networks (DNNs) has been shown to scale with the size of both the training dataset and model architecture~\citep{hestness2017deep}; however, the resources required to deploy large networks for inference can be prohibitive as they often exceed the \icc{compute and storage} budgets of mobile or edge devices~\citep{n8_han2015deep, n38_gale2019state}.
To minimize inference costs, quantization and pruning are widely used techniques that reduce \icc{resource} requirements by respectively limiting the precision of and removing elements from DNNs.
}

{In this work, we propose a framework \icc{for} \ianUpdate{jointly} applying novel methods for uniform quantization and unstructured pruning to both the features and weights of DNNs \ianUpdate{during training}\footnote{The code for the algorithms introduced in this paper can be found at \ourlib.}.}
\ian{The majority of state-of-the-art techniques for quantization-aware training calculate gradients using the straight-through estimator (STE), which is notoriously sensitive to weight initialization~\citep{yin2019understanding, gholami2021survey}.}
\ian{To account for this, we propose a modification we refer to as \textit{delayed quantization}, in which we postpone the STE-based calculations \ianUpdate{to later training \xinyu{epochs}.}}
\ian{When \xinyu{applied} to generative adversarial networks (GANs) trained on image-to-image translation tasks, we observe a long-tailed distribution \icc{of activations,} similar to the distribution of weights observed by~\cite{n23_wang2019qgan}.}
\Question{\ian{To minimize the impact of outliers, we introduce another modification we refer to as \textit{saturated quantization}, in which we clip the long-tailed distribution of {activations} based on quantiles determined during training.}}
\ian{Finally, we extend the unstructured weight \ianUpdate{pruning} technique proposed by~\cite{n45_zhu2017prune} to the activation space.}
\ian{To {the best of} our knowledge, we are the first to thoroughly evaluate the impact of unstructured \ianUpdate{activation} pruning.}

\ian{Quantization and pruning techniques are often considered to be independent problems~\citep{n24_paupamah2020quantisation, n21_liang2021pruning};
however, recent work has begun to study the \ianUpdate{joint} application of both in a unified scope~\citep{n10_zhao2019focused, n17_yu2020joint,n4_van2020bayesian}.}
In this work, we aim to more deeply understand the relationship between these optimizations and, in doing so, address two key \ianUpdate{issues overlooked in previous work}: (1) quantization and pruning techniques are often analyzed for either discriminative or generative tasks, rarely both; (2) frameworks for joint quantization and pruning currently default to the ``prune-then-quantize” \icc{training schedule} \ianUpdate{depicted in Fig.~\ref{fig:prune_then_quantize}} without exploring the alternative.
\Question{Thus, we evaluate our {framework} across a wide range of {both} discriminative and generative tasks and  {consider the ``quantize-the-prune" training schedule in addition to the standard paradigm}.}
\Question{{In doing so,} we observe a non-commutative \xinyu{relationship} when applying our novel quantization and pruning methods to both the weights and activations of DNNs.}
\ian{Based on these results, we state \textit{the non-commutativity hypothesis.}} \\

\noindent \textbf{The Non-Commutativity Hypothesis.} \ian{\textit{For a given deep neural network being trained for a specific task, there exists an exact training schedule in which pruning and quantization can be introduced to optimize \icc{the performance of the network on that task.}}} \\

\icc{We empirically evaluate this hypothesis and demonstrate that the optimal ordering in which quantization and pruning can be introduced into the training schedule not only exists but also varies across discriminative and generative tasks.
Thus, our results advocate a rethinking of the currently accepted \quotes{prune-then-quantize} paradigm.}
\Question{Using the optimal training schedules determined within our framework, we show increased performance per memory footprint over existing solutions.} \\

\textbf{We summarize our contributions as follows:}

\aistatCond{\vspace{-0.5cm}}{}
\begin{enumerate}
    \item We propose a framework to train deep neural networks using \ian{novel methods for} uniform quantization and unstructured pruning on both the weights and activations \ian{(Section~\ref{sec:method})}.

    \item \ian{We demonstrate the non-commutative nature of of quantization and pruning when applied to the weights and activations of DNNs (Section~\ref{sec:non_commutativity_hypothesis}).}
	      	      	      
	\item \ian{We show that our method delivers the best network performance per memory footprint across a wide range of discriminative and generative computer vision tasks when compared to existing state-of-the-art solutions (Section~\ref{sec:comparisons})}.
\end{enumerate}


\begin{figure*}[]
	\centering
	\subfloat[]{\includegraphics[width=0.5\linewidth]{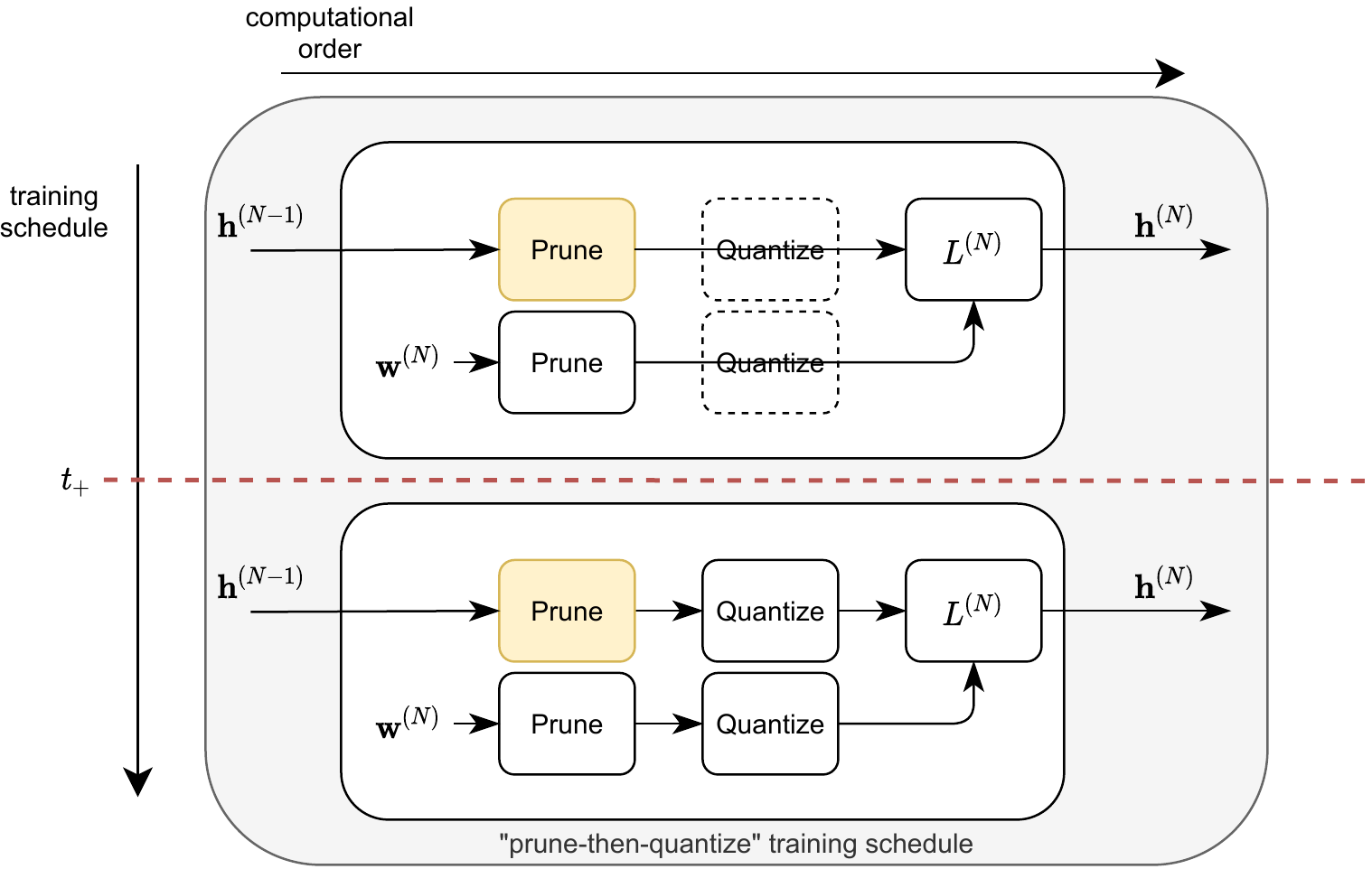}}
	~~
	\subfloat[]{\includegraphics[width=0.5\linewidth]{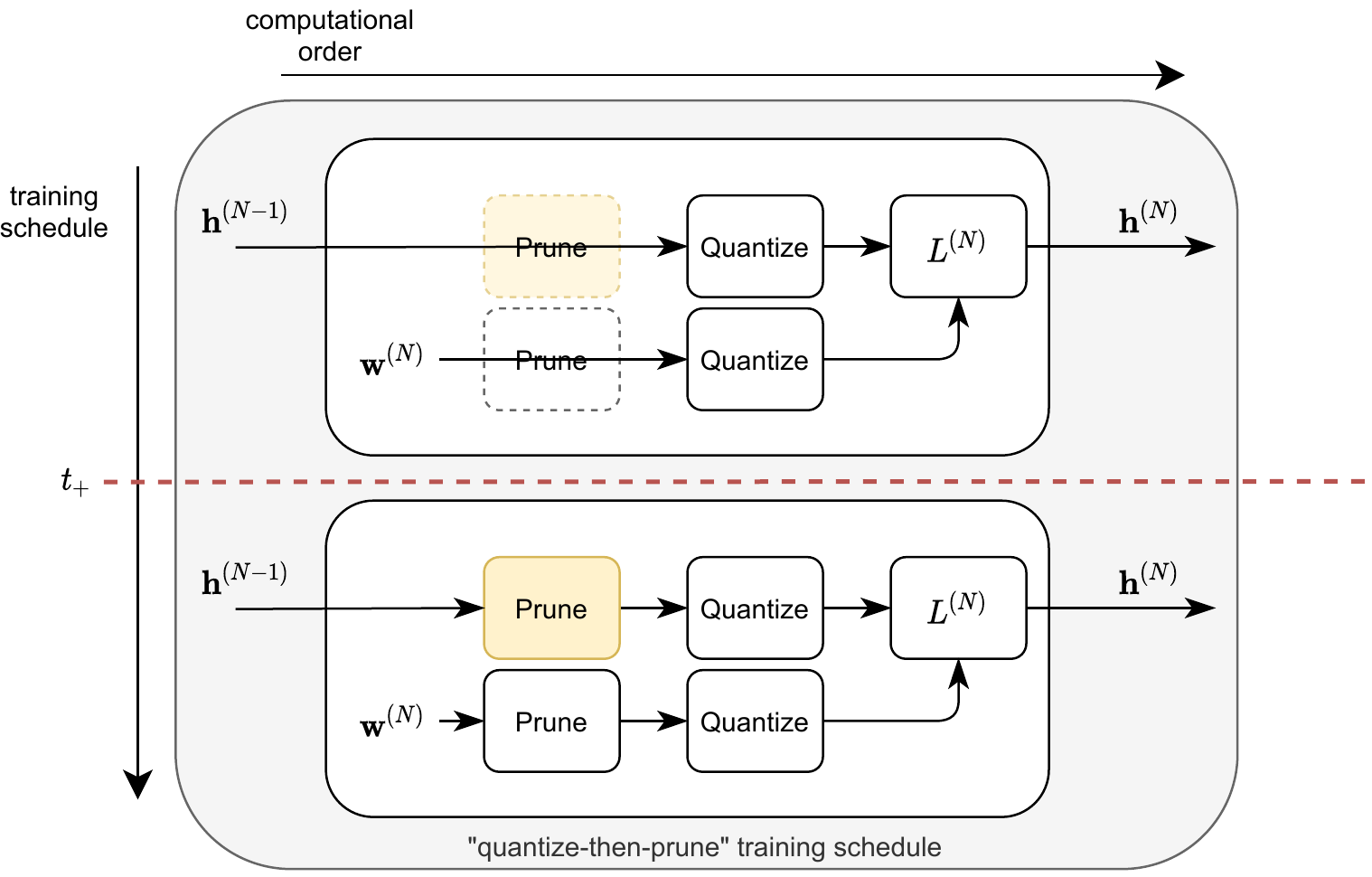}}
    \caption{\small{\icc{As discussed in Section~\ref{sec:method}, our framework differs from existing solutions in to key ways:}
    \ianUpdate{(1) we apply unstructured pruning to the activations of the network (\textbf{\textcolor{BurntOrange}{yellow}}); and
    (2) \icc{we enable the consideration of both the standard \quotes{prune-then-quantize} training schedule (left) as well as its \quotes{quantize-then-prune} analog (right).}
    \icc{Using this framework,} we empirically evaluate \textit{the non-commutativity hypothesis} \icc{over both discriminative and generative tasks, as discussed in Section~\ref{sec:empirical_analysis}.}}
    \Question{Note that under both paradigms, the latent operator \xinyu{(dotted)} is activated at time $t_+$, at which point both quantization and pruning are applied to both the weights and activations of the DNN.}
    }}
	\label{fig:pipeline}
\end{figure*}

\aistatCond{\vspace{-0.3cm}}{} 

\section{Background}
\label{sec:pre}

\ianUpdate{Here, we provide background on the quantization and pruning techniques further detailed in Section~\ref{sec:method}.}





\subsection{\ianUpdate{Uniform} Quantization}
\label{sec:pre_quantize}

{Quantization minimizes the inference costs by reducing the precision requirements of deep neural networks (DNNs).}
\ianUpdate{By \textit{uniform} quantization, we refer to the process of reducing the precision of all weights and activations to the same number of bits.}
{With energy consumption dominated by data movement, using lower precision representations reduces the energy consumed when running inference by decreasing the memory footprint required to load and store weights and activations while increasing the speed of computations~\citep{horowitz20141, n28_krishnamoorthi2018quantizing}.}
{It has been shown that DNNs are resilient \ianUpdate{to} these reductions in precision, and this resilience has been exploited to deploy state-of-the-art models on resource-constrained hardware~\citep{n25_jacob2018quantization,n16_wu2020integer,n4_van2020bayesian}.}

{Approaches to quantization are often divided into two categories: post-training quantization and quantization-aware training.}
{Whereas the former applies quantization after a network has been trained, the latter directly considers the loss due to quantization error throughout the training process.}
{{It has been shown that} quantization-aware training often outperforms post-training quantization ~\citep{n28_krishnamoorthi2018quantizing}.}
{In Section~\ref{sec:my_quantization}, we \ianUpdate{introduce} novel methods for quantization-aware training that \ianUpdate{uniformly limit the precision of} both the weights and activations of DNNs.}
	
\subsection{Unstructured Pruning}
\label{sec:pre_prune}

\ianUpdate{Pruning is the process of identifying redundant elements to be removed from the computation graph of a DNN.}
\ianUpdate{In practice, this is often done by setting the values of the identified elements to zero.}
\ianUpdate{The proportion of zero-valued elements is referred to as the network's \textit{sparsity}, where a higher value corresponds to fewer non-zero elements~\citep{n38_gale2019state}.}

\ianUpdate{Pruning techniques are often divided into \textit{unstructured} or \textit{structured} approaches which define if and how to impose a pre-defined topology over the computation graph~\citep{n38_gale2019state}.}
\ianUpdate{While structured pruning techniques (\textit{e.g.}, neuron- or channel-pruning) offer simple implementations that efficiently \icc{map to} modern GPUs, unstructured pruning techniques offer higher compression rates due to their inherent flexibility~\citep{n13_gray2017gpu, n29_liu2018rethinking, n35_hoefler2021sparsity}.}

\ianUpdate{Numerous works have studied the benefits of applying unstructured pruning techniques to the weights of DNNs \icc{throughout training}~\citep{n45_zhu2017prune, n37_frankle2018lottery, n29_liu2018rethinking, n38_gale2019state, n35_hoefler2021sparsity}; however, the benefits of \icc{also} applying unstructured pruning to the activations have yet to be fully explored in the current literature.
\icc{Existing work \xinyu{on pruning activations} has exploited the inherent sparsity patterns in the activations of DNNs to guide post-training \icc{model compression techniques}~\citep{hu2016network} and studied the impact of imposing \textit{structured} pruning on the activations of DNNs during training~\citep{wang2020structured}.}
\icc{However, it has been shown that applying \textit{unstructured} weight pruning during training achieves higher compression rates while maintaining network performance when compared to post-training or structured techniques~\citep{n29_liu2018rethinking, n38_gale2019state}}.}
\icc{In this work, we are interested in {introducing} sparsity patterns to guide \textit{unstructured} pruning of both the weights and activations of DNNs throughout training.}
\icc{To do so, we extend the unstructured pruning techniques proposed by ~\cite{n45_zhu2017prune} to the activations of DNNs to yield even higher compression rates while maintaining network performance, as further discussed in Section~\ref{sec:my_pruning}.}


\aistatCond{\vspace{-0.3cm}}{}
\section{\ianUpdate{Proposed \icc{Framework}}}
\label{sec:method}


\ianUpdate{
As depicted in Fig.~\ref{fig:pipeline}, our framework differs from existing solutions in two key ways:
    (1) we jointly apply \icc{our} methods for uniform quantization and unstructured pruning to both the weights and activations of DNNs during training; and
    (2) \icc{we enable the consideration of both the standard \quotes{prune-then-quantize} training schedule as well as its \quotes{quantize-then-prune} analog.}}
\icc{As the converse of the \quotes{prune-then-quantize} training schedule, the \quotes{quantize-then-prune} paradigm deactivates the pruning operator until time $t_+$, when it is then introduced into the computational order throughout training.}
\ianUpdate{Here, we describe the core methods of our framework.}

\subsection{\ianUpdate{Uniform} Quantization Method}
\label{sec:my_quantization}

\ianUpdate{As shown in} Eq.~\ref{eq:uniform_q}, \ianUpdate{we denote the} uniform quantization \ianUpdate{operator} as $Q_u(\mat{x}, d)$, where $\mat{x}$ denotes the input to the operator (\textit{i.e.}, weights or activations), $N$ denotes the total number of bits used \ianUpdate{to represent weights and activations}, and $d$ \ianUpdate{denotes} the number of bits used to represent \ianUpdate{the fractional bits} \ian{to the right of the decimal}.

\aistatCond{\vspace{-0.5cm}}{}
\begin{equation}
	Q_u(\mat{x}, d) = \text{clip}(\floor{\mat{x} \times 2^{d}}, -2^{N-1}, 2^{N-1}-1) / 2^d
	\label{eq:uniform_q}
\end{equation}

\ianUpdate{To calculate gradients, we use the standard straight-through estimator (STE) given by Eq.~\ref{eq:ste} which is shown to have superior convergence properties~\citep{hinton2012neural, o_hubara2017quantized}.}
\Question{As summarized in Section~\ref{sec:intro}, we introduce delayed and saturated quantization techniques to stabilize STE-based quantization-aware training across both discriminative and generative tasks.}

\aistatCond{\vspace{-0.3cm}}{}
{\small
\begin{equation}
	\frac{\partial Loss}{\partial \mat{x}} = \text{clip}(\frac{\partial Loss}{\partial Q_u(\mat{x}, d)}, -2^{N-d-1}, 2^{N-d-1} - 2^{-d})
	\label{eq:ste}
\end{equation}}

\aistatCond{\vspace{-0.3cm}}{}
\noindent \textbf{Delayed Quantization.}
\ianUpdate{To account for the instability observed in STE-based quantization-aware training techniques, we delay the quantization of the network to later training stages.}
Starting with the original full-precision network, \ianUpdate{we} calculate the optimal decimal bits ($d^*$) by minimizing the quantization error after a given number of \ianUpdate{update} steps ($t_q$), as shown in Eq.~\ref{eq:optimal_decimal_bits}.
\ianUpdate{Here, $\mat{x}_t$ denotes the weights or activations at time $t$.}
\ianUpdate{Eq.~\ref{eq:delayed} shows the resulting delayed quantization operator\footnote{In this work, we focus on uniform quantization; however, $Q_u(\mat{x}, d)$ can be \xinyu{also} replaced with a mixed-precision variant.}.}

\aistatCond{\vspace{-0.3cm}}{}
\begin{equation}
d^* = \arg \min_{d} \Vert Q_u(\mat{x}_{t_q}, d) - \mat{x}_{t_q} \Vert^2
\label{eq:optimal_decimal_bits}
\end{equation}

\aistatCond{\vspace{-0.5cm}}{}
\begin{equation}
	\small
	\begin{split}
	Q_{u,D}(\mat{x}_t) & = \begin{cases} 
		\mat{x}_t & t < t_q \\
		Q_u(\mat{x}_t, d^*)  &    t \ge t_q   \\
	 \end{cases} \\
	 \end{split}
	\label{eq:delayed}
\end{equation}

\begin{table}
	\footnotesize
	\centering
	\resizebox{20em}{!}{
	\begin{tabular}{|l|l|}
		\hline
		\textbf{Quantization Method}                             & \textbf{Top-1 Acc} \\ \hline
		Baseline                                        &  92.60              \\
		\hline
		$Q_{u,D}(\mat{x_t})$                                         &  92.52            \\ \hline
		$Q_u(\mat{x}_t, N-1)$   & 75.83               \\ 
		\hline
		$Q_u(\mat{x}_t, \hat{d}^*)$                                    & 18.35
		\\ \hline
	\end{tabular}
	}
    \caption{\small{\ianUpdate{We train MobileNetV2 on the CIFAR10 database at 8-bit precision to analyze the performance of our delayed quantization technique $Q_{u,D}(\mat{x_t})$ against alternate uniform quantization strategies. Here, the baseline is the full precision network. $Q_u(\mat{x_t}, N-1)$ follows the approach of \cite{n25_jacob2018quantization} and $Q_u(\mat{x_t}, \hat{d}^*)$ trains the network from scratch using $d^*$ as determined by $Q_{u,D}(\mat{x_t})$.}}}
	\label{tb:delayed_benefit}
\end{table}	

\begin{table*}[t!]
	\begin{minipage}[t]{0.24\textwidth}
		\footnotesize
		\resizebox{\textwidth}{!}{
			\begin{tabular}{|l|c|}
				\hline
				\textbf{{\scriptsize (a) MobileNet}}          & \textbf{{ 
				Acc}}  \\ \hline
				Baseline             & 92.60       \\ \hline
				\q{w,f} & 92.52   \\ \hline
			\end{tabular}
		}
	\end{minipage}
	\begin{minipage}[t]{0.24\textwidth}
		\footnotesize
		\resizebox{\textwidth}{!}{
			\begin{tabular}{|l|c|}
				\hline
				\textbf{\scriptsize (b) ResNet101}          & \textbf{mAP}  \\ \hline
				Baseline             & 74.47       \\ \hline
				\q{w,f} & 74.25  \\ \hline
			\end{tabular}
		}
	\end{minipage}
	\begin{minipage}[t]{0.24\textwidth}
		\footnotesize
		\resizebox{\textwidth}{!}{
			\begin{tabular}{|l|c|}
				\hline
				\textbf{\small (c) ESPCN}          & \textbf{PSNR}  \\ \hline
				Baseline             & 32.84        \\ \hline
				\q{w,f} & 32.68   \\ \hline
			\end{tabular}
		}
	\end{minipage}
	\begin{minipage}[t]{0.24\textwidth}
		\footnotesize
		\resizebox{\textwidth}{!}{
			\begin{tabular}{|l|c|}
				\hline
				\textbf{\small (d) Pix2Pix}          & \textbf{FID}  \\ \hline
				Baseline             & 119.90        \\ \hline
				\q{w,f} & 118.50   \\ \hline
			\end{tabular}
		}
	\end{minipage}
    \aistatCond{\vspace{-1em}}{}
    \caption{\small{\ianUpdate{We evaluate our delayed quantization method across a variety of network architectures trained \icc{over a} wide range of \icc{either} discriminative \icc{or} generative tasks.}
    Experiment settings and notation are \icc{further detailed} in Section~\ref{sec:empirical_analysis}. }}
	\label{tb:delayed_all}
	\aistatCond{\vspace{-1em}}{}
\end{table*}

\ianUpdate{To demonstrate the benefits of our approach, we compare our delayed quantization operator $Q_{u,D}(\mat{x_t})$ against two alternate uniform quantization strategies when training MobileNetV2~\citep{mobilenetv2} on CIFAR10~\citep{o_cifar10} at 8-bit precision\footnote{We detail the configurations used for our delayed quantization experiments in the appendix.}.}
\ianUpdate{First, we compare against \cite{n25_jacob2018quantization}, who propose to set the \xinyu{number of} decimal bits to $N-1$; we denote this as $Q_u(\mat{x_t}, N-1)$.}
\ianUpdate{Second, we consider the case in which we use the optimal decimal bits ($d^*$) resulting from $Q_{u,D}(\mat{x_t})$ to retrain a new network from scratch; we denote this as $Q_u(\mat{x}_t, \hat{d}^*)$.}
\ianUpdate{The results are given in Table~\ref{tb:delayed_benefit}, where the full precision network is provided as a baseline.}
In all experiments, we use the standard Xavior initialization, as is common practice~\citep{glorot2010understanding}.

It can be clearly seen that our method \ianUpdate{minimizes performance loss due to quantization error when compared to alternative strategies}.
\ianUpdate{While both} $Q_u(\mat{x}_t, N-1)$ and $Q_u(\mat{x}_t, \hat{d}^*)$ apply \ianUpdate{constant} STE-based quantization \ianUpdate{throughout} training, $Q_{u,D}(\mat{x_t})$ \ianUpdate{adaptively fits to distribution shifts caused by DNN parameter updates.}
\ianUpdate{We believe the large performance difference between $Q_{u,D}(\mat{x_t})$ and $Q_u(\mat{x}_t, \hat{d}^*)$ indicates that the distribution of weights and activations of DNNs undergo drastic shifts during training that are not easy to recover from without the adaptive strategy employed by our delayed quantization technique.}
Unlike previous works~\citep{n25_jacob2018quantization, o_init_q}, which use floating point auxiliary weights for simulated quantization or find clever weight initializations to stabilize performance, we argue \ianUpdate{that} our method is simpler, more \ianUpdate{efficiently implemented}, and more robust.

\Question{Table~\ref{tb:delayed_all} provides a thorough evaluation of our delayed quantization method applied across a variety of network architectures to uniformly reduce network precision to 8 bits.
{Over} this wide range of discriminative \icc{and} generative tasks, we show that our training method minimizes performance loss due to quantization error when compared to their full precision counterparts.} \\

\begin{figure}[]
    \centering
    \subfloat[Activations]{\includegraphics[width=0.4\linewidth]{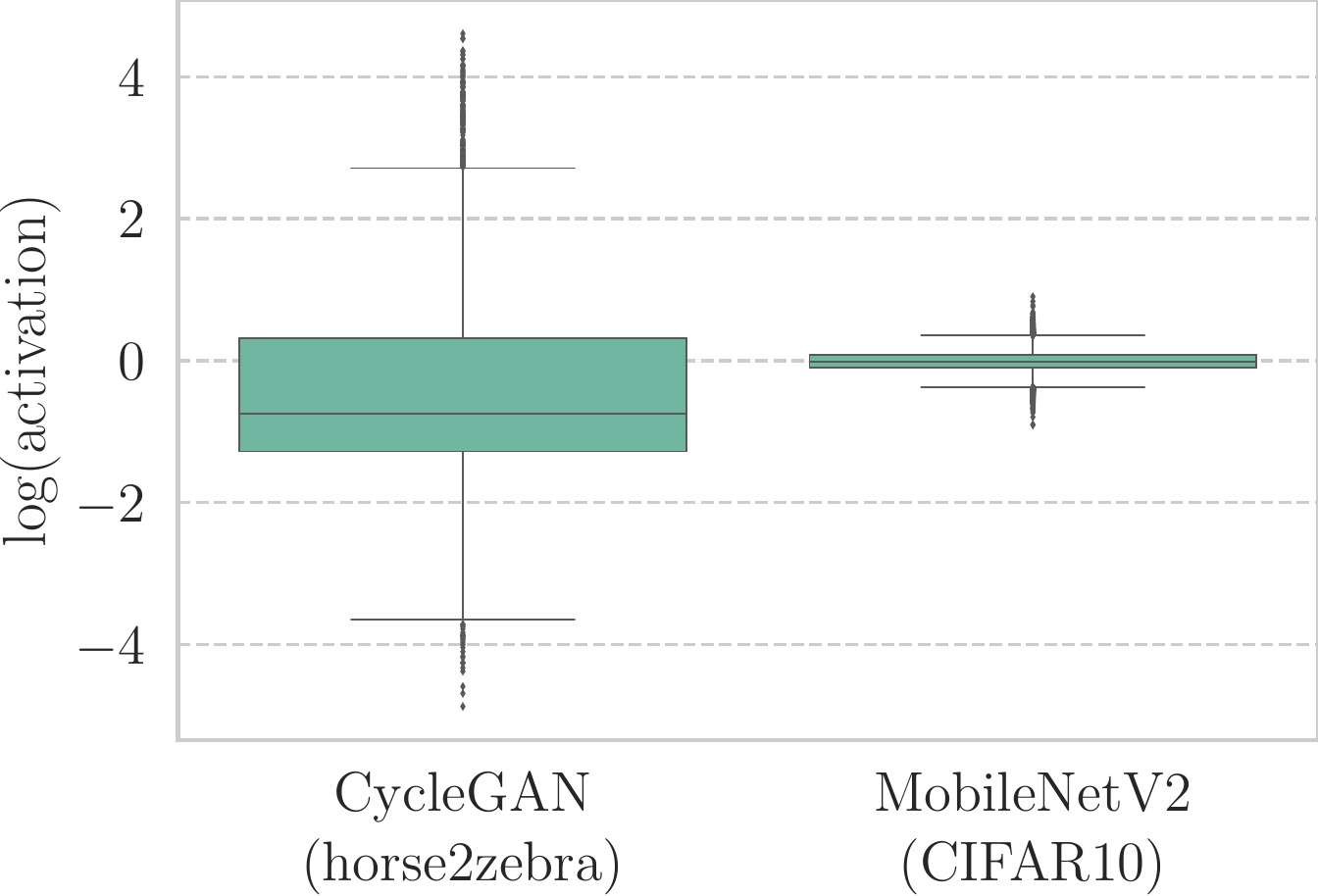}}
    ~~
    \subfloat[Weights]{\includegraphics[width=0.4\linewidth]{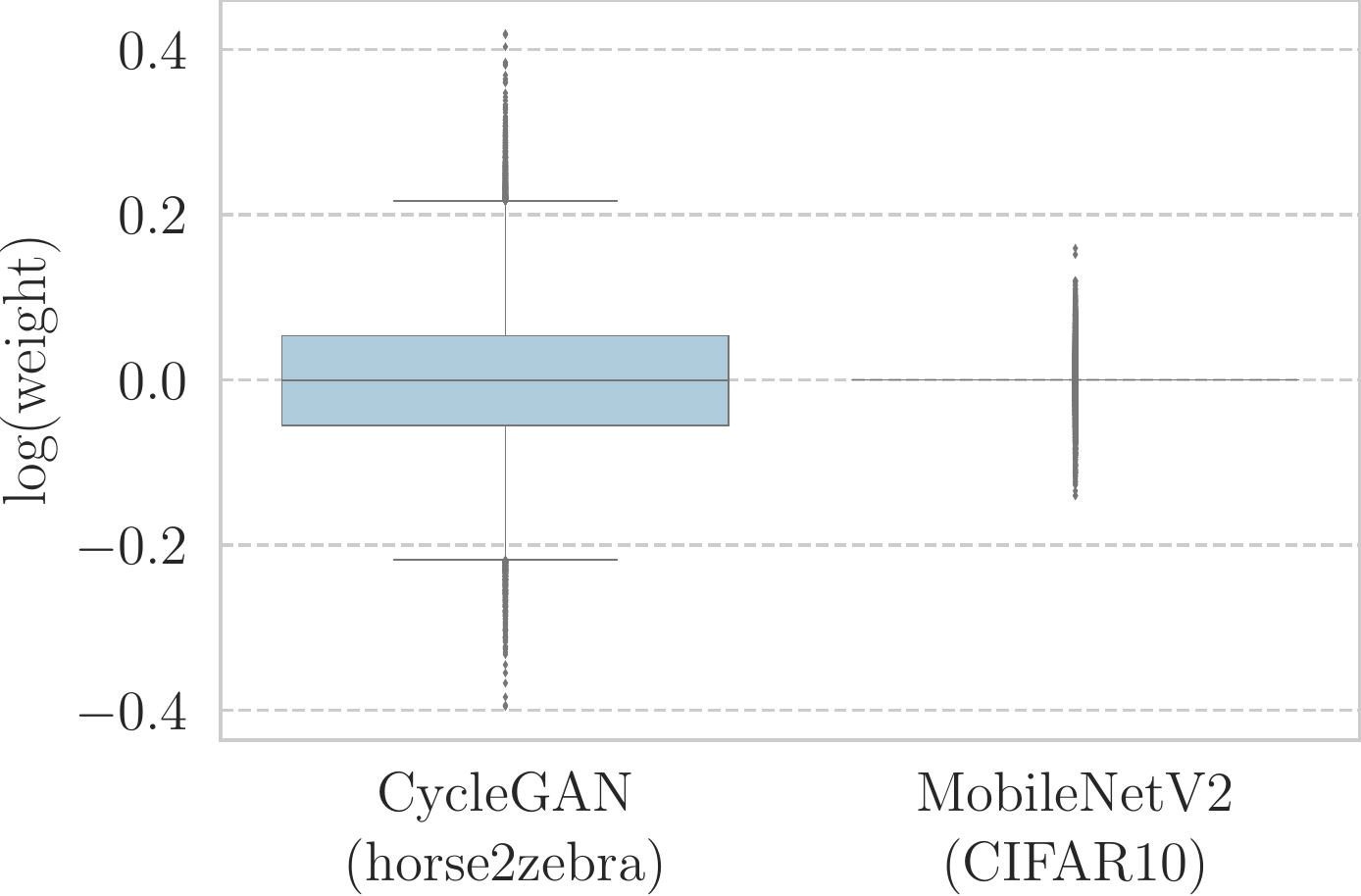}}
\caption{\small{\ianUpdate{Here, we compare the distribution of activations sampled from CycleGAN to those sampled from MobileNetV2 to underscore the extreme outliers that impact delayed quantization. Experiment settings and notations are further detailed in Section~\ref{sec:empirical_analysis}.}}}
\label{fig:feat_dist}
\end{figure}

\begin{figure}[]
    \centering
    \subfloat[]{\includegraphics[width=0.20\linewidth]{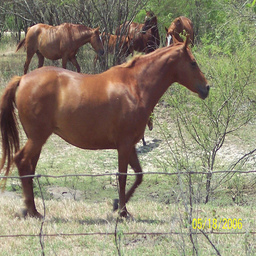}}
    ~~
    \subfloat[]{\includegraphics[width=0.20\linewidth]{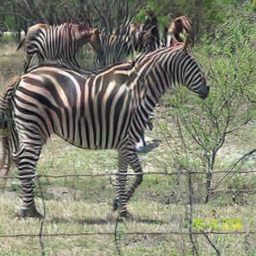}}
    ~~
    \subfloat[]{\includegraphics[width=0.20\linewidth]{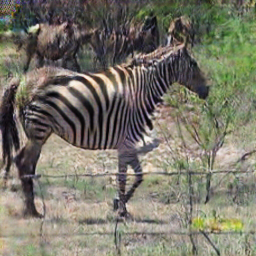}}
    ~~
    \subfloat[]{\includegraphics[width=0.20\linewidth]{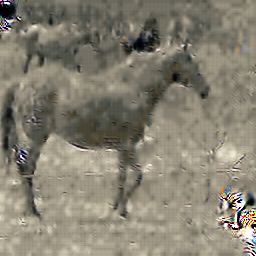}}
\caption{\small{\ianUpdate{These images were generated from CycleGAN trained on ``horse2zebra" using (a) as the input. The baseline (b) shows the resulting translation at full precision, (c) shows the resulting translation when trained at 8-bit precision with saturated quantization over the activation space as well as 50\% weight and activation pruning, and (d) shows the resulting translation without saturated quantization. Experiment settings and notations are further detailed in Section~\ref{sec:empirical_analysis}.}}}
\label{fig:cyclegan_saturation}
\end{figure}

\noindent \textbf{Saturated Quantization.} 
\ianUpdate{Due to the adaptive nature of our delayed quantization operator, the estimation of the optimal decimal bits ($d^*$) as given by Eq.~\ref{eq:optimal_decimal_bits} can be sensitive to extreme outliers in the weight or activation \xinyu{distributions}.}
\ianUpdate{In our experiments, we observe a long-tailed distribution only in the activation space of DNNs trained on generative tasks --- more specifically, generative adversarial networks (GANs). }
\ian{As depicted} in Fig.~\ref{fig:feat_dist}, the distribution of activations sampled from CycleGAN~ \citep{o_cyclegan} \ianUpdate{trained on ``horse2zebra" has a much wider range} than that of MobileNetV2 trained on CIFAR10.
These \ianUpdate{extreme outliers} result in the underestimation of $d^*$ and \ianUpdate{lead to catastrophic degradation of image fidelity, as shown in Fig.~\ref{fig:cyclegan_saturation}.}

\ianUpdate{\cite{n23_wang2019qgan} observe a similar phenomenon in the weights space of GANs and attribute this reduction in network performance to the under-representation of quantized values.}
Unlike their solution, which is a GAN-specific quantization method \ianUpdate{based on} the \ianUpdate{expectation-maximization (EM)} algorithm, we discover that \ianUpdate{this} under-representation issue can be simply and effectively moderated by clipping the outliers.

Before \ianUpdate{estimating the optimal decimal bits ($d^*$)} using Eq.~\ref{eq:optimal_decimal_bits}, we apply a data-driven saturation function $S_{q_l, q_u}(\mat{x})$ defined by Eq.~\ref{eq:saturate}.
\ianUpdate{Here,} $\mat{x}$ denotes the input to the operator (\textit{e.g.}, activations), $q_l$ denotes the lower quantile of the distribution of sampled activations using Eq.~\ref{eq:quantile}, and $q_u$ similarly denotes the upper quantile.
\ianUpdate{Note that $q_l$ and $q_u$ are hyperparameters such that $q_l, q_u \in [0, 1]$ and  $q_l < q_u$.
In our experiments with CycleGAN, we use quantiles $q_l=0.01\%$ and $q_u=99.99\%$.}

\aistatCond{\vspace{-0.5cm}}{}
\begin{equation}
	S_{q_l, q_u}(\mat{x}) = \text{clip}(\mat{x}, \small{\text{quantile}}(\mat{x}, q_l), \small{\text{quantile}}(\mat{x}, q_u)) 
	\label{eq:saturate}
\end{equation}
\aistatCond{\vspace{-0.3cm}}{}
\begin{equation}
	\text{quantile}(\mat{x}, a) =  a\text{-th quantile of } \mat{x}
	\label{eq:quantile}
\end{equation}

\aistatCond{\vspace{-0.1cm}}{}

\ianUpdate{Thus, when using delayed quantization for GANs, we estimate the optimal decimal bits using Eq.~\ref{eq:optimal_decimal_bits_saturdates}.}
As shown in  Fig.~\ref{fig:cyclegan_saturation}, \ianUpdate{applying saturated quantization to the activations of GANs preserves} generative quality.

\aistatCond{\vspace{-0.3cm}}{}
\begin{equation}
d^* = \arg \min_{d} \Vert Q_u(\mat{x}_{t_q}, d) - S_{q_l, q_u}(\mat{x}_{t_q}) \Vert^2
\label{eq:optimal_decimal_bits_saturdates}
\end{equation}


\aistatCond{\vspace{-0.3cm}}{}
\subsection{\ianUpdate{Unstructured} Pruning Method}
\label{sec:my_pruning}

\ianUpdate{Magnitude-based unstructured weight pruning has been shown to yield impressive compression rates while maintaining network performance similar to their fully connected counterparts~\citep{n8_han2015deep, n45_zhu2017prune, n38_gale2019state}.}
\ianUpdate{To do so, this class of techniques} use weight magnitude as a \ianUpdate{surrogate for} importance when \ianUpdate{identifying redundant weights to remove from the computation graph}.
\cite{n45_zhu2017prune} propose an unstructured pruning technique that maintains a binary mask for each set of weights and allows for the reactivation of pruned elements throughout training.
\ianUpdate{In this work, we extend this technique to also prune the activations of deep neural networks (DNNs).}

\ianUpdate{As shown in Eq.~\ref{eq:prune}, we denote the unstructured pruning operator $P(\mat{x}, s)$ \xinyu{as \ianLast{element-wise} multiplication between $\mat{x}$ and $\mat{M}_{\mat{x},s}$}, where $\mat{x}$ denotes the input to the operator (\textit{i.e.}, weights or activations), $s$ denotes the \ian{target sparsity as measured by} the \ian{percentage} of zero\ian{-valued elements}, and $\mat{M}_{\mat{x},s}$ denotes its binary mask.}
Given that $(i,j)$ are the row and column indices\ianUpdate{, respectively,} the binary mask $\mat{M}_{\mat{x},s}$ is calculated using Eq.~\ref{eq:binary_mask} where the quantile operation is defined by Eq.~\ref{eq:quantile}.


\aistatCond{\vspace{-1.5em}}{}

\begin{equation}
	P(\mat{x}, s)  = \mat{x} \circ \mat{M}_{\mat{x},s}
	\label{eq:prune}
\end{equation}

\aistatCond{\vspace{-2.5em}}{}

\begin{equation}
	\begin{split}
	\mat{M}_{\mat{x},s}^{(i,j)} & = \begin{cases}
		1 & |\mat{x}^{(i, j)}| \ge \text{quantile}(|\mat{x}|, s) \\
		0 & \text{otherwise}
	\end{cases}
	\end{split}
	\label{eq:binary_mask}
\end{equation}

\aistatCond{\vspace{-0.3cm}}{}
\ianUpdate{As proposed by \cite{n45_zhu2017prune}, the} sparsity \ianUpdate{level} ($s$) is controlled and updated \ianUpdate{according to} a sparsification schedule at time steps $t_p + i \Delta t_p$ \ianUpdate{such that} $i \in \{1,2,..,,n\}$, where $t_p$, $\Delta t_p$, and $n$ are hyperparameters
that represent the starting step, frequency, and \ianUpdate{total number of steps}, respectively.


\begin{figure}
	\centering
    \includegraphics[width=0.6\linewidth]{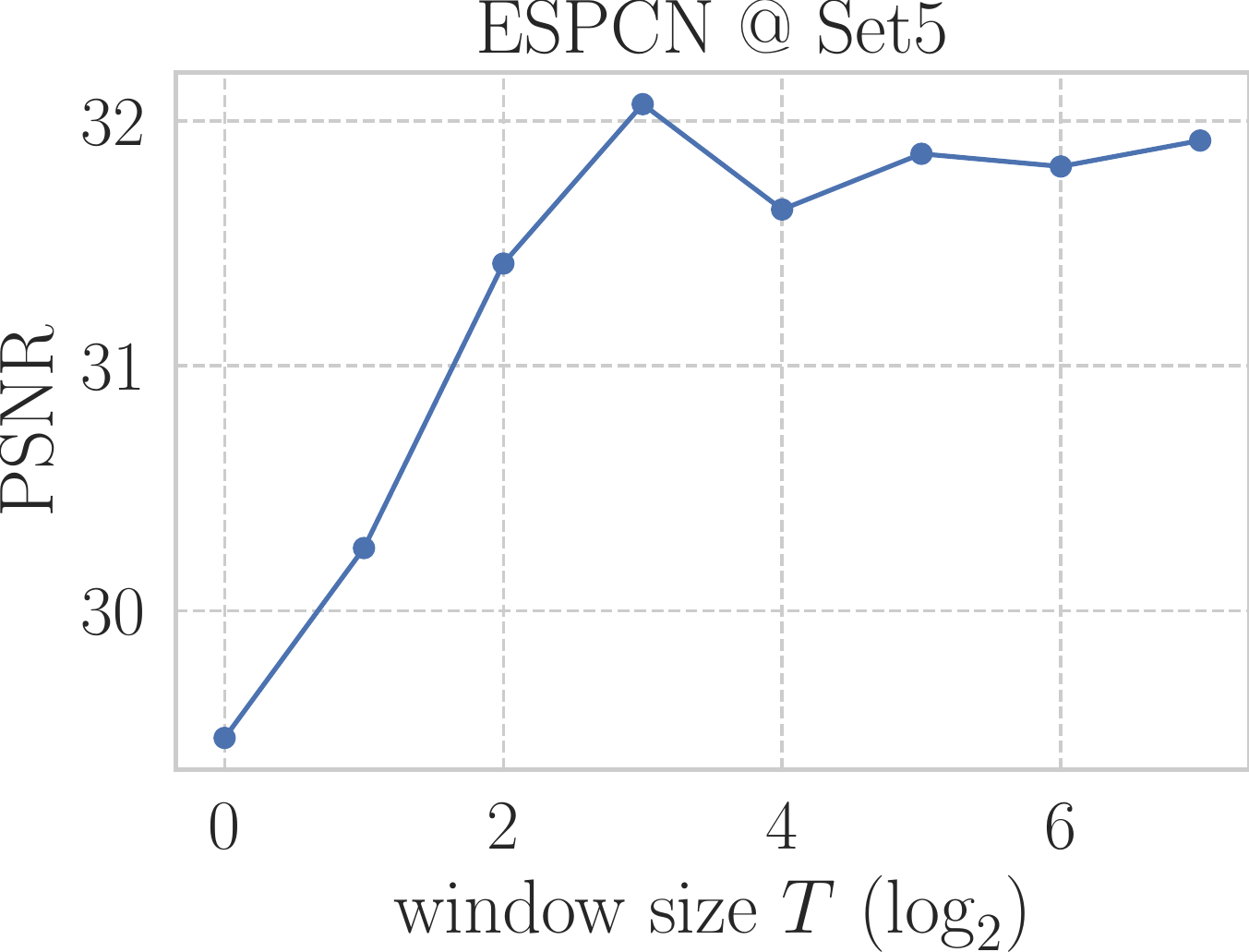}
    \caption{\small{\ianUpdate{Here, we evaluate the effect of window size ($T$) on the resulting PSNR of the ESPCN super resolution network evaluated over Set5.}}}
	\label{fig:window_size}
\end{figure}

\ianUpdate{As an artifact of the commonly used rectified linear unit (ReLU), the activations of DNNs are inherently sparse~\citep{glorot2011deep, xu2015empirical}; however, this sparsity pattern is neither structured nor static.}
Contrary to the weights of DNNs, which \textit{are} static during inference, \ianUpdate{the activations of DNNs are dynamically conditioned on the input to the network~\citep{zhou2015cnnlocalization}.}
\ianUpdate{\icc{This dynamic sparsity pattern} is difficult to exploit or optimize during inference.}
\ianUpdate{To account for this} without sacrificing \icc{computational} efficiency, we introduce a sliding window \ianUpdate{technique to stabilize} unstructured \ianUpdate{activation} pruning \icc{throughout training.}

Let $\mat{h}_t$ denote the \ianUpdate{activations} at time $t$ and $T$ denote the size of the sliding window.
\ianUpdate{We extend the calculation of the binary mask to the activations of DNNs as given by Eq.~\ref{eq:binary_mask} by evaluating the mask over a running sum within a sliding window of size $T$ as is shown in Eq.~\ref{eq:binary_mask_feature}.}

\aistatCond{\vspace{-0.3cm}}{}
{\footnotesize
\begin{equation}
	\mat{M}_{\mat{h}_t,s}(i,j) = \begin{cases}
		1 & \sum\limits_{n=0}^{T-1} |\mat{h}_{t-n}(i, j)| \ge \text{\scriptsize quantile}(\sum\limits_{n=0}^{T-1} |\mat{h}_{t-n}|, s) \\
		0 & \text{otherwise}
	\end{cases}  
	\label{eq:binary_mask_feature}
\end{equation}}

\aistatCond{\vspace{-0.3cm}}{}
In Fig.~\ref{fig:window_size}, we evaluate the effect of window size ($T$) \ianUpdate{on the} 
PNSR of a super resolution network (ESPCN) \citep{espcn} on Set5 \citep{o_set5}.
\ianUpdate{We observe that while PSNR increases with $T$, the benefit of an increased window size saturates around 16.}
\ianUpdate{It is important to note that this not only} indicates that our approach can estimate the expected activation pattern, it also aligns with the observation of \cite{hanin2019deep} that ReLU-based DNNs \ianUpdate{express} much fewer activation patterns than their theoretical limit.
\ianUpdate{By applying unstructured pruning to the activation space of these DNNs, we exploit} the severe under-utilization of \ianUpdate{a network's} potential expressiveness \ianUpdate{and significantly increase compression rates while maintaining performance}.

\section{Empirical Analysis}
\label{sec:empirical_analysis}

\begin{table*}[]
	\begin{minipage}[t]{0.33\textwidth}
		\resizebox{\textwidth}{!}{
			\begin{tabular}{|l|c|c|}
				\hline
				\textbf{\small (a) Image Classification}    & \textbf{Top-1 Acc} & \textbf{PD} \\ \hline
				Baseline           & 92.60   & 0.31         \\ \hline \hline
				\pq{w}{w,f} & \textbf{92.23}   & \textbf{1.43}    \\ \hline
				\qp{w,f}{w} & 85.94   & 1.33   \\ \hline
			\end{tabular}
		}
	\end{minipage}
	\begin{minipage}[t]{0.33\textwidth}
		\resizebox{\textwidth}{!}{
			\footnotesize
			\begin{tabular}{|l|c|c|}
				\hline
				\textbf{\small (b) Object Detection}    & \textbf{mAP} & \textbf{PD}  \\ \hline
				Baseline              & 74.47  & 0.33       \\ \hline \hline
				\pq{w}{w,f} & \textbf{74.44}  & \textbf{0.71} \\ \hline
				\qp{w,f}{w} & 74.13   & 0.71 \\ \hline
			\end{tabular}
		}
	\end{minipage}
	\begin{minipage}[t]{0.33\textwidth}
		\footnotesize
		\resizebox{\textwidth}{!}{
			\begin{tabular}{|l|c|c|}
				\hline
				\textbf{\small (c) Super Resolution}          & \textbf{PSNR} & \textbf{PD} \\ \hline
				Baseline             & 32.84   & 6.22      \\ \hline \hline
				\pq{w}{w,f} & 32.51  & 8.36 \\ \hline
				\qp{w,f}{w} & \textbf{32.54}  & \textbf{8.37} \\ \hline
			\end{tabular}
		}
	\end{minipage}
    \caption{\small{\ianUpdate{Here, we empirically evaluate the \textit{non-commutativity hypothesis} across both discriminative and generative computer vision tasks. While experiments (a) and (b) favor the standard ``prune-the-quantize" training schedule, experiment (c) indicates the need to consider the alternative. Note that notations and experiment settings are discussed in Section~\ref{sec:empirical_analysis}.}}}
	\label{tb:evaluate_previous}
\end{table*}

\begin{table*}[]
	\begin{minipage}[t]{0.33\textwidth}
		\resizebox{\textwidth}{!}{
			\begin{tabular}{|l|c|c|}
				\hline
				\textbf{\small (a) Image Classification}    & \textbf{Top-1 Acc} & \textbf{PD} \\ \hline
				Baseline           & 92.60    & 0.31          \\ \hline \hline
				\pq{w,f}{w,f} & \textbf{91.44}  & \textbf{2.49}    \\ \hline
				\qp{w,f}{w,f} & 86.84   & 2.36    \\ \hline
			\end{tabular}
		}
	\end{minipage}
	\begin{minipage}[t]{0.33\textwidth}
		\resizebox{\textwidth}{!}{
			\footnotesize
			\begin{tabular}{|l|c|c|}
				\hline
				\textbf{\small (b) Object Detection}    & \textbf{mAP} & \textbf{PD}  \\ \hline
				Baseline              & 74.47  & 0.33       \\ \hline \hline
				\pq{w,f}{w,f} & \textbf{73.00} & \textbf{0.86} \\ \hline
				\qp{w,f}{w,f} & 70.13  & 0.82 \\ \hline
			\end{tabular}
		}
	\end{minipage}
	\begin{minipage}[t]{0.33\textwidth}
		\footnotesize
		\resizebox{\textwidth}{!}{
			\begin{tabular}{|l|c|c|}
				\hline
				\textbf{\small (c) Super Resolution}          & \textbf{PSNR} & \textbf{PD} \\ \hline
				Baseline             & 32.84   & 6.22      \\ \hline \hline
				\pq{w,f}{w,f} & 31.03   & 8.34 \\ \hline
				\qp{w,f}{w,f} & \textbf{31.66}   & \textbf{8.51} \\ \hline
			\end{tabular}
		}
	\end{minipage}
    \caption{\small{\ianUpdate{Here, we extend the experiments summarized in Table~\ref{tb:evaluate_previous} to include the unstructured activation pruning techniques introduced in Section~\ref{sec:my_pruning}. Doing so further accentuates the exact ordering in which quantization and pruning should be introduced into the training schedule to optimize network performance.}}}
	\label{tb:non-comu}
\end{table*}

In this section, we empirically evaluate the \textit{non-commutativity hypothesis} and compare our framework to existing state-of-the-art solutions.
\ianUpdate{We demonstrate that our methods deliver superior performance per memory footprint across the following discriminative and generative} computer vision tasks:
\aistatCond{\vspace{-0.2cm}}{}
\begin{enumerate}[(a)]
    \item \textbf{Image classification} with MobileNetV2 \citep{mobilenetv2} on CIFAR10 \citep{o_cifar10}
    \item \textbf{Object detection} with Faster RCNN \citep{o_fasterrcnn} using ResNet101 \citep{he2016deep}
on Pascal VOC \citep{o_pascalvoc}
    \item \textbf{Super resolution} with ESPCN
\citep{espcn} on Set5 \citep{o_set5}
    \item \textbf{Image-to-image translation} with two generative adversarial networks (GANs): Pix2Pix~\citep{pix2pix} and CycleGAN~\citep{o_cyclegan} trained on facades and horse2zebra, respectively
\end{enumerate}

We implement each task using the existing open-source implementations \ianUpdate{as provided by their respective authors.}
We maintain \ianUpdate{their selected} weight initialization \ianUpdate{techniques} and \ianUpdate{apply all of the same hyperparameters aside from the number of training epochs}.
We \ianUpdate{provide further} details \ianUpdate{on} experiment setups in \xinyu{Appendix~\ref{apn:detailed_exp}}.

\Question{{As discussed in Section~\ref{sec:method}, our framework enables the consideration of} both the \quotes{prune-then-quantize} and \quotes{quantize-then-prune} training schedules when applying our novel uniform quantization and unstructured pruning methods to both the weights and activations of DNNs.
{Given quantization delay $t_q$ and pruning delay $t_p$, we define $t_+ = \text{max}(t_p, t_q)$ following Fig.~\ref{fig:pipeline}.}
{Note that this results in the \quotes{quantize-the-prune} paradigm when $t_q < t_p$ and \quotes{prune-then-quantize} otherwise.}}

\ianUpdate{We denote the ``prune-then-quantize" training schedule as \pq{w,f}{w,f} when applying both quantization and pruning to both the weights and activations of a DNN using 8-bit precision and a target sparsity of 50\%.}
\ianUpdate{We denote its ``quantize-then-prune" analog as \qp{w,f}{w,f}.}
\ianUpdate{Similarly, we denote the standard ``prune-then-quantize" training schedule as \pq{w}{w,f} for the case where no pruning is applied to the activations of the DNN.}

\ianUpdate{Finally,} we introduce a metric \ian{we refer to as} \textit{performance density} (PD), which we define as the domain-specific performance (\textit{e.g.}, accuracy, PSNR, mAP) divided by the resulting memory footprint of the combined weights and activations.
\ianUpdate{Because increased levels of quantization and pruning invariably lead to drops in network performance, we use this metric to analyze the trade-off between compression and performance across both discriminative and generative tasks.}

\subsection{Non-Commutativity Hypothesis}
\label{sec:non_commutativity_hypothesis}

\ianUpdate{To evaluate the non-commutativity hypothesis, we start by evaluating the currently accepted ``prune-then-quantize" training schedule \pq{w}{w,f} across experiments (a), (b), and (c).}
\ianUpdate{We compare this training schedule to its ``quantize-then-prune" analog \qp{w,f}{w} and provide the results in Table~\ref{tb:evaluate_previous}.}


\ianUpdate{Our results indicate a need to rethink the commonly accepted ``prune-then-quantize" paradigm.}
\ianUpdate{Whereas experiments (a) and (b) support the standard training schedule, experiment (c) highlights the benefits of considering the alternative.}
\ianUpdate{Our hypothesis is further strengthened when we apply our unstructured activation pruning methods detailed in Section~\ref{sec:my_pruning}.}

\ianUpdate{As shown in Table~\ref{tb:non-comu}, applying unstructured pruning to the activations in both training schedules further reinforces the correct order that optimizes performance density (PD).}
When comparing Tables~\ref{tb:evaluate_previous} and \ref{tb:non-comu}, 
it \ianUpdate{is observed} that \ianUpdate{the addition of activation} pruning \ianUpdate{universally} provides higher PD \ianUpdate{as the compression of the activation space significantly reduces the total memory footprint while maintaining network performance.}

\ianUpdate{This result not only empirically validates} that there exists an optimal order \ianUpdate{in which} quantization and pruning \ianUpdate{should be introduced into the training schedule to optimize network performance, it also indicates that} this ordering varies \ianUpdate{across discriminative and generative} tasks.
\ianUpdate{As shown in Table~\ref{tb:inception_score}, this trend holds when extended to experiment (d)---GANs trained for image-to-image translation.}
\ian{The majority of} existing works \ian{studying} quantization and pruning \ian{techniques primarily} focus on discriminative tasks \ian{(\textit{e.g.}, image classification) rather than} generative tasks \ian{(\textit{e.g.},} image-to-image translation).
\ian{\xinyu{Recent} studies have shown impressive results when applying these techniques to GANs~\citep{n23_wang2019qgan, n34_zhou2020sparse}.}
\ianUpdate{We extend our framework to Pix2Pix (d-1) and CycleGAN (d-2) and observe that both networks favor the ``\ianLast{quantize}-then-prune" training schedule\footnote{
 We provide sample images in the appendix.}.}
{\footnotesize
\begin{table}[]
\footnotesize
	\centering
		\begin{table}[H]
		\centering
		\footnotesize
			\begin{tabular}{|l|l|l|}
			\hline
			\textbf{(d-1) Pix2Pix} & \textbf{FID} & \textbf{PD}  \\ \hline
			Baseline         &  119.9     & 3.67         \\ \hline \hline
			\pq{w,f}{w,f}    &  154.8      & 22.37       \\ \hline
			\qp{w,f}{w,f}    &  {135.0}   & \textbf{25.66}       \\ \hline
			\end{tabular}
		\end{table}
        \aistatCond{\vspace{-0.8cm}}{}
			\begin{table}[H]
			    \centering
			    \footnotesize
				\begin{tabular}{|l|l|l|}
				\hline
				\textbf{(d-2) CycleGAN} & \textbf{FID} & \textbf{PD} \\ \hline
				Baseline         & 67.1         & 3.28   \\ \hline \hline
				$P_{0.5}(w,f)\rightarrow Q_{12}(w,f)$    & 100.4   & 16.67        \\
				\hline
				$Q_{12}(w,f) \rightarrow P_{0.5}(w,f)$      & {89.4}   & \textbf{18.73}       \\
				\hline
				\end{tabular}
			\end{table}
    \aistatCond{\vspace{-1em}}{}
    \caption{\small{\ianUpdate{Here, we extend the experiments summarized in Table~\ref{tb:non-comu} to GANs trained for image-to-image translation. To quantify generative quality, we use the inverse of the Fréchet Inception Distance (FID)~\citep{heusel2017gans} between the generated samples and real images as the domain-specific performance metric used to calculate PD. Note that a lower FID indicates higher generative quality. In our experiments, we find the CycleGAN activations require a higher bitwidth to maintain network performance.}}}
	\label{tb:inception_score}
	\aistatCond{\vspace{-1em}}{}
\end{table}
}

\begin{table*}[t!]
	\centering
	\resizebox{\textwidth}{!}{
		\begin{tabular}{ccccccccccccc}
			\toprule
			Method
			& Network                                           & $N_{W}$ & $N_A$                   & $s_W$                  & $s_A$                & {\footnotesize Baseline Acc}                            &
			\multicolumn{1}{c}{Accuracy}  & \multicolumn{1}{c}{\footnotesize
			\begin{tabular}[c]{c}Weights (Mb) \end{tabular}} &
			\multicolumn{1}{c}{\footnotesize \begin{tabular}[c]{c}Activations (Mb)
			\end{tabular}} &
			\multicolumn{1}{c}{\footnotesize \begin{tabular}[c]{c}PD (Acc/Mb)
			\end{tabular}}   \\
			\midrule
			\multicolumn{1}{c|}{\script{\citep{n40_choi2016towards}}}
			& \multicolumn{1}{c|}{ResNet-32}                    &
			\multicolumn{1}{c|}{8}    & \multicolumn{1}{c|}{--}  &
			\multicolumn{1}{c|}{77.8\%} & \multicolumn{1}{c|}{--}   &
			\multicolumn{1}{c|}{92.58}                 & 92.64 (+0.06)    &
			37.80 & 131.07 & 0.55			\\ \midrule
			\multicolumn{1}{c|}{\script{\citep{n44_achterhold2018variational}}}
			& \multicolumn{1}{c|}{DenseNet-76}                  &
			\multicolumn{1}{c|}{2}    & \multicolumn{1}{c|}{--}  &
			\multicolumn{1}{c|}{54\%}   & \multicolumn{1}{c|}{--}   &
			\multicolumn{1}{c|}{92.19}                 & 91.17 (-1.02)  &
			\textbf{0.68} & 282.53 & 0.32                 \\ \midrule
			
			\multicolumn{1}{c|}{\multirow{3}{*}{\script{\citep{n29_liu2018rethinking}}}}
			& \multicolumn{1}{c|}{VGG-19}                       &
			\multicolumn{1}{c|}{--}   & \multicolumn{1}{c|}{--}  &
			\multicolumn{1}{c|}{80\%}   & \multicolumn{1}{c|}{--}   &
			\multicolumn{1}{c|}{93.5}                  & 93.52 (+0.02)       &
			128.26 & 38.80 &  0.56          \\
			\multicolumn{1}{c|}{}
			& \multicolumn{1}{c|}{PreResNet-110}                &
			\multicolumn{1}{c|}{--}   & \multicolumn{1}{c|}{--}  &
			\multicolumn{1}{c|}{30\%}   & \multicolumn{1}{c|}{--}   &
			\multicolumn{1}{c|}{95.04}                 & 95.06 (+0.02) &
			952.03 & 619.71  & 0.06             \\
			\multicolumn{1}{c|}{}
			& \multicolumn{1}{c|}{DenseNet-100}                 &
			\multicolumn{1}{c|}{--}   & \multicolumn{1}{c|}{--}  &
			\multicolumn{1}{c|}{30\%}   & \multicolumn{1}{c|}{--}   &
			\multicolumn{1}{c|}{95.24}                 & \textbf{95.21} (-0.03)  &
			27.11 & 426.74 & 0.21                \\ \midrule
			
			\multicolumn{1}{c|}{\script{\citep{n43_zhao2019variational}}} & \multicolumn{1}{c|}{DenseNet-40} & \multicolumn{1}{c|}{--} & \multicolumn{1}{c|}{--} & \multicolumn{1}{c|}{59.7\%} & \multicolumn{1}{c|}{--} & \multicolumn{1}{c|}{94.11} 
			& 93.16 (-0.95)    & 3.23 & 114.87 & 0.79               \\ \midrule

			\multicolumn{1}{c|}{\script{\citep{n3_xiao2019autoprune}}}
			& \multicolumn{1}{c|}{VGG-16}                       &
			\multicolumn{1}{c|}{--}   & \multicolumn{1}{c|}{--}  &
			\multicolumn{1}{c|}{79\%}   & \multicolumn{1}{c|}{--}   &
			\multicolumn{1}{c|}{93.40}                  & 91.5 (-1.9)        &
			98.97 & 35.39  &  0.68		\\ \midrule

			\multicolumn{1}{c|}{\multirow{2}{*}{\script{\citep{n33_dettmers2019sparse}}}}
			& \multicolumn{1}{c|}{VGG16-C}                      &
			\multicolumn{1}{c|}{--}   & \multicolumn{1}{c|}{--}  &
			\multicolumn{1}{c|}{95\%}   & \multicolumn{1}{c|}{--}   &
			\multicolumn{1}{c|}{93.51}                 & 93 (-0.51)   &  23.57 & 35.39 & 1.58                  \\
			\multicolumn{1}{c|}{}
			& \multicolumn{1}{c|}{WRN-22-8}                     &
			\multicolumn{1}{c|}{--}   & \multicolumn{1}{c|}{--}  &
			\multicolumn{1}{c|}{95\%}   & \multicolumn{1}{c|}{--}   &
			\multicolumn{1}{c|}{95.74}                 & 95.07 (-0.67)     &
			27.46 & 230.69  & 0.37		\\ \midrule

			\multicolumn{1}{c|}{\script{\citep{n2_yang2020automatic}}}
			& \multicolumn{1}{c|}{ResNet-20}                    &
			\multicolumn{1}{c|}{1.9}  & \multicolumn{1}{c|}{--}  &
			\multicolumn{1}{c|}{54\%}   & \multicolumn{1}{c|}{--}   &
			\multicolumn{1}{c|}{91.29}                 & 91.15 (-0.14)  & 9.77 & 78.64 &  1.03  \\ \midrule

			\multicolumn{1}{c|}{\script{\citep{n4_van2020bayesian}} }
			& \multicolumn{1}{c|}{VGG-7}                        &
			\multicolumn{1}{c|}{4.8} & \multicolumn{1}{c|}{5.4} &
			\multicolumn{1}{c|}{--}    & \multicolumn{1}{c|}{--}   &
			\multicolumn{1}{c|}{93.05}                 & 93.23 (+0.18)
			&  43.85 & \textbf{3.27} &  1.98		\\ \midrule
			\multicolumn{1}{c|}{\script{\citep{n24_paupamah2020quantisation}}}
			& \multicolumn{1}{c|}{MobileNet}                    &
			\multicolumn{1}{c|}{8}    & \multicolumn{1}{c|}{8}   &
			\multicolumn{1}{c|}{--} & \multicolumn{1}{c|}{--}   &
			\multicolumn{1}{c|}{91.31}                 & 90.59 (-0.72)        &
			25.74 & 13.17 & 2.33		\\ \midrule

			\multicolumn{1}{c|}{\script{\citep{n41_choi2020universal}}}
			& \multicolumn{1}{c|}{ResNet-32}                    &
			\multicolumn{1}{c|}{8}  & \multicolumn{1}{c|}{--}  &
			\multicolumn{1}{c|}{87.5\%}  & \multicolumn{1}{c|}{--}   &
			\multicolumn{1}{c|}{92.58}                 & 92.57 (-0.01)    &
			21.28 & 131.07 &  0.61			\\ \midrule
			
			\multicolumn{1}{c|}{\multirow{2}{*}{Ours}}
			& \multicolumn{1}{c|}{\multirow{2}{*}{MobileNetV2}} &
			\multicolumn{1}{c|}{8}    & \multicolumn{1}{c|}{8}   &
			\multicolumn{1}{c|}{50\%}     & \multicolumn{1}{c|}{--}   &
			\multicolumn{1}{c|}{\multirow{2}{*}{92.60}} &
			\multicolumn{1}{c}{92.23 (-0.37)} & 9.19 & 55.20  & 1.43 \\
			\multicolumn{1}{c|}{}
			& \multicolumn{1}{c|}{}                             &
			\multicolumn{1}{c|}{8}    & \multicolumn{1}{c|}{8}   &
			\multicolumn{1}{c|}{50\%}     & \multicolumn{1}{c|}{50\%} &
			\multicolumn{1}{c|}{}                      &
			\multicolumn{1}{c}{91.44 (-1.16)} &  {9.19} & {27.60}  & \textbf{2.49}
			\\ \bottomrule
		\end{tabular}
	} 
    \caption{\small{\ianUpdate{We demonstrate the superior performance per memory footprint (\textit{i.e.}, performance density) of our framework when compared to existing image classification solutions trained on CIFAR10.
    By jointly applying uniform quantization and unstructured pruning to both the weights and activations of the DNN, we achieve a higher performance density (PD) with lower compression rates than existing solutions and comparable network performance.}}}
	\label{tb:comparisons}
\end{table*}





\subsection{Comparing Against Existing Methods}
\label{sec:comparisons}

\ianUpdate{Using the optimal orderings determined in Section~\ref{sec:non_commutativity_hypothesis}, we demonstrate that our framework achieves superior performance per memory footprint (\textit{i.e.} performance density) when compared to existing image classification solutions trained on on CIFAR10.}
\ianUpdate{Table~\ref{tb:comparisons} summarizes this comparison, where $N_W$ and  $N_A$} denote the average number of bits used to quantize the weights or activations, respectively.
\ianUpdate{Similarly,} $s_W$ and $s_A$ denote the average weight and activation sparsity, respectively. 

From Table~\ref{tb:comparisons}, \ianUpdate{it can be observed} that our method is \ianUpdate{uniquely} comprehensive, supporting both quantization and pruning over both \ianUpdate{the weights
and activations.}
\ianUpdate{Furthermore,} our \ianUpdate{framework} achieves the smallest memory footprint and highest
performance density (PD) \ianUpdate{when applying unstructured pruning to the activation space.}
Unlike methods such as \citep{n2_yang2020automatic, n4_van2020bayesian}, our framework provides direct control over the target bitwidth and sparsity, which is more useful in practical scenarios \ianUpdate{under tight design constraints}.

\aistatCond{\vspace{-0.2cm}}{}
\section{Discussion}
\aistatCond{\vspace{-0.8em}}{}
\label{sec:discussion}

\aistatCond{\vspace{-0.1cm}}{}
\icc{It has been shown that activations dominate data movement and therefore time and energy costs during inference in real-time computer vision applications~\citep{horowitz20141, jha120data, h_colbert2021energy}.}
\icc{Extending unstructured pruning to the activation space not only significantly improves the network performance per memory footprint, it also reduces the cost of running inference in resource-constrained settings with limited compute and memory budgets such as edge or mobile computing.}

\ianUpdate{The prevalence of deep learning solutions for edge and mobile applications has resulted in a trend of minimizing the compute and memory requirements of over-parameterized networks.} \ianUpdate{However, since the performance of DNNs is known to scale with the size of the network~\citep{hestness2017deep}, using PD to compare existing solutions that apply quantization and pruning across various neural network architectures provides a holistic view of the compression-accuracy trade-off.}

\ianUpdate{When analyzing our framework,} Tables~\ref{tb:non-comu} and \ref{tb:inception_score} show discriminative tasks favor the \quotes{prune-then-quantize} training schedule while generative tasks favor the alternative.
Based on these results, we articulate the \textit{non-commutativity conjecture.} \\

\noindent \textbf{The Non-Commutativity Conjecture.} \textit{The optimal schedule in which quantization and pruning are introduced throughout training is intimately tied to the magnitude of the network's gradient updates.} \\

\ianUpdate{Applying} unstructured activation pruning \ianUpdate{throughout training invariably} results in shifts in the distribution of activations, which are known to lead to exploding gradients~\citep{littwin2018regularizing}.
\ianUpdate{However, applying} STE-based quantization \ianUpdate{inherently constrains both} the activation distribution and their resulting gradients using clipping as shown in Eq.~\ref{eq:uniform_q} and~\ref{eq:ste}.
Additionally, it has been shown that batch normalization stablilizes the distribution of gradients~\citep{santurkar2018does}.
Whereas the authors of the selected discriminative tasks (a) and (b) both utilize batch normalization, the authors of the selected generative tasks (c) and (d) do not.
\ianUpdate{Thus, we conjecture that for the selected generative tasks, quantization acts as a regularizer to stabilize the distribution shifts caused by unstructured activation pruning.}

\section{Conclusions and Future Work}

\aistatCond{\vspace{-1em}}{}

We propose a framework \ianUpdate{to apply} novel methods for uniform quantization and
unstructured pruning to both \ianUpdate{the weights and activations of DNNs during training.}
\ianUpdate{To the best of our knowledge, we are the first to} thoroughly evaluate \ianUpdate{the} impact of applying unstructured \ianUpdate{activation} pruning \icc{in this setting.}
\icc{Unlike previous work, our framework enables the consideration of both the standard \quotes{prune-then-quantize} training schedule as well as its \quotes{quantize-the-prune} analog.}
\icc{Using this framework, we evaluate the performance of our methods when jointly applied to DNNs trained over a wide range of discriminative and generative tasks.}
\icc{Based on our observations, we articulate and evaluate \textit{the non-commutativity hypothesis} to determine that the optimal ordering in which quantization and pruning are introduced into the training schedule not only exists, but also varies across tasks.}
\ianUpdate{Using the optimal training schedules for each task, we demonstrate the superior performance per memory footprint of our framework when compared to existing solutions.}
In future work, we aim to extend our analyses to larger network architectures trained on more complex datasets for tasks not limited to computer vision.
\ianUpdate{Additionally, we aim to explore the application of neural architecture search algorithms in automating the selection and design of layer-specific training schedules.}




\bibliographystyle{apalike}

\clearpage

\appendix

\onecolumn

{\noindent\LARGE \textbf{Appendix}}

\section{Software Library}
\label{apn:software}

\ianUpdate{The code for each algorithm introduced in this paper can be found in our Python library, \texttt{qsparse},  at \ourlib.} In Listing~\ref{code:interface}, we \ianUpdate{provide an example of} how to use our algorithms for uniform quantization and unstructured pruning as PyTorch modules~\citep{paszke2019pytorch}.

The majority of existing model compression and quantization Python libraries provide a minimal set of \ianUpdate{specialized} modules \ianUpdate{separated}, which limits the flexibility of the software interface~\citep{t_xilinx, coelho2020ultra, QatTorch}.
\ianUpdate{However, the simplicity of our framework enables an efficient software interface.}
\ianUpdate{In contrast, \texttt{qsparse}} supports a wide range of neural network specifications.
To improve the library flexibility, we \ianUpdate{introduce} a technique which directly transforms the weight attribute of the input layer into a pruned or quantized version at runtime.
Thus, our library is layer-agnostic and can work with any PyTorch module as long as their parameters can be accessed from their weight attribute, as is \ianUpdate{standard practice}~\citep{paszke2019pytorch}.

\begin{listing}[H]
\begin{Verbatim}[commandchars=\\\{\}]
\PYG{k+kn}{import} \PYG{n+nn}{torch.nn} \PYG{k}{as} \PYG{n+nn}{nn}
\PYG{k+kn}{from} \PYG{n+nn}{qsparse} \PYG{k+kn}{import} \PYG{n}{prune}\PYG{p}{,} \PYG{n}{quantize}

\PYG{c+c1}{\PYGZsh{} feature pruning and quantization}
\PYG{n}{nn}\PYG{o}{.}\PYG{n}{Sequential}\PYG{p}{(}
	\PYG{n}{nn}\PYG{o}{.}\PYG{n}{Conv2d}\PYG{p}{(}\PYG{l+m+mi}{10}\PYG{p}{,} \PYG{l+m+mi}{30}\PYG{p}{,} \PYG{l+m+mi}{3}\PYG{p}{),}
	\PYG{n}{prune}\PYG{p}{(}\PYG{l+m+mf}{0.5}\PYG{p}{,} \PYG{n}{start}\PYG{o}{=}\PYG{l+m+mi}{1000}\PYG{p}{,} \PYG{n}{interval}\PYG{o}{=}\PYG{l+m+mi}{100}\PYG{p}{,} \PYG{n}{repetition}\PYG{o}{=}\PYG{l+m+mi}{3}\PYG{p}{),} 
	\PYG{n}{quantize}\PYG{p}{(}\PYG{n}{bits}\PYG{o}{=}\PYG{l+m+mi}{8}\PYG{p}{,} \PYG{n}{timeout}\PYG{o}{=}\PYG{l+m+mi}{2000}\PYG{p}{)}  \PYG{c+c1}{\PYGZsh{} `timeout` is `quantize step` in Alg. 1}
\PYG{p}{)}                                 

\PYG{c+c1}{\PYGZsh{} weight pruning and quantization (layers other than `Conv2d` work as well)}
\PYG{n}{qpconv} \PYG{o}{=} \PYG{n}{quantize}\PYG{p}{(}\PYG{n}{prune}\PYG{p}{(}\PYG{n}{nn}\PYG{o}{.}\PYG{n}{Conv2d}\PYG{p}{(}\PYG{l+m+mi}{10}\PYG{p}{,} \PYG{l+m+mi}{30}\PYG{p}{,} \PYG{l+m+mi}{3}\PYG{p}{),} \PYG{l+m+mf}{0.5}\PYG{p}{),} \PYG{l+m+mi}{8}\PYG{p}{)}
\end{Verbatim}

	\caption{Examples of our software interface for quantization and pruning on both weights and features}
	\label{code:interface}
\end{listing}

\section{Experiment Setup}
\label{apn:detailed_exp}

\ianUpdate{Using our library, we introduce minimal code changes to the original open-source implementations used for our experiments.}
\xinyu{We will release the code of each experiment as examples for our library.}
\ianUpdate{For each experiment,} we \ianUpdate{maintain the} hyperparameters \ianUpdate{selected by the} original \ianUpdate{authors aside from the number of} training epochs.
\ianUpdate{Table~\ref{tbl:hyperparams} summarizes the hyperparameters considered in our experiments.}

\begin{table}[H]
\centering
\begin{tabular}{cl}
\hline
\multicolumn{1}{c}{\textbf{Param}}    & \textbf{Description}                  \\ \hline \hline
\multicolumn{1}{|c|}{$t_q$}         & \multicolumn{1}{l|}{The number of epochs to delay quantization} \\ \hline
\multicolumn{1}{|c|}{$q_u$} & \multicolumn{1}{l|}{The upper quantile used for saturated quantization} \\ \hline
\multicolumn{1}{|c|}{$q_l$} & \multicolumn{1}{l|}{The lower quantile used for saturated quantization} \\ \hline
\multicolumn{1}{|c|}{$t_p$} & \multicolumn{1}{l|}{The number of epochs to delay unstructured pruning} \\ \hline
\multicolumn{1}{|c|}{$\Delta t_p$} & \multicolumn{1}{l|}{The frequency of updates to the binary mask} \\ \hline
\multicolumn{1}{|c|}{$n$} & \multicolumn{1}{l|}{The total number of pruning steps applied throughout training} \\ \hline
\multicolumn{1}{|c|}{$T$} & \multicolumn{1}{l|}{The window size used for unstructured activation pruning} \\ \hline
\multicolumn{1}{|c|}{$E$} & \multicolumn{1}{l|}{The total number of training epochs} \\ \hline
\end{tabular}
\caption{\ianUpdate{Summary of the hyperparameters considered in this work}}
\label{tbl:hyperparams}
\end{table}

\subsection{Image Classification on Cifar10} 
\label{apn:cifar10}

\ianUpdate{We use~\citep{kuangliu13/pytorch-cifar} as the foundation of our image classification experiments and train MobileNetV2~\citep{mobilenetv2} on CIFAR10~\citep{o_cifar10}.}
We quantize all the weights and activations to 8-bit precision and use a target sparsity of 50\%.
{We do not quantize the bias as it is typically stored and computed at full precision during on-device inference~\citep{n25_jacob2018quantization}.}
We apply unstructured pruning to both the weights and activations of all hidden layers, \ianUpdate{leaving the first and last layers densely connected}.
We set the total number of training epochs ($E$) to 250, $q_u$ to 0, $q_l$ to 1, and $T$ to
$2048$ in each experiment.  

\begin{table}[H]
\centering
\begin{tabular}{cl}
\hline
\multicolumn{1}{c}{\textbf{Experiment}}    & \textbf{Settings}                  \\ \hline \hline
\multicolumn{1}{|c|}{\pq{*}{*}}         & \multicolumn{1}{l|}{
\begin{tabular}[l]{l}
$t_q$ was set to 230 and 235 for the weights and activations, \\
respectively. $t_p, \Delta t_p,\text{ and } n$ were set to $100,15, \text{ and }4$, respectively.
\end{tabular}
} \\ \hline
\multicolumn{1}{|c|}{\qp{*}{*}} & \multicolumn{1}{l|}{
\begin{tabular}[l]{l}
$t_q$ was set to 160 and 170 for the weights and activations, \\
respectively. $t_p, \Delta t_p,\text{ and } n$ were set to 180,15, and 4, respectively.
\end{tabular}
} \\ \hline
\multicolumn{1}{|c|}{\q{*}} & \multicolumn{1}{l|}{
\begin{tabular}[l]{l}
$t_q$ was set to 230 and 240 for the weights and activations, respectively.
\end{tabular}
} \\ \hline
\end{tabular}
\caption{\ianUpdate{Summary of the hyperparameters used for image classification}}
\label{tbl:image_classification}
\end{table}

\ianUpdate{We calculate the {performance density} (PD) for this task as the accuracy divided by the combined size of the weights and activation of the network in megabits (Mb), such that the units are accuracy-per-megabit (Acc/Mb).
The memory footprint of the activations is calculated assuming an input size of (3, 32, 32).}

\subsection{Super Resolution on Set5} 

\ianUpdate{We use~\citep{yjn870/ESPCN-pytorch} as the foundation of our super resolution experiments and train ESPCN~\citep{espcn} for using 91-image \citep{o_91image}.}
\ianUpdate{Similar to the experiments outlined in Appendix~\ref{apn:cifar10}, we quantize all weights and activations at 8-bit precision using a target sparsity of 50\% without pruning for first and last layer of the network.}
\ianUpdate{We set $E$ to 200, $q_u$ to 0, $q_l$ to 1, and $T$ to $16$ in each experiment.}

\begin{table}[H]
\centering
\begin{tabular}{cl}
\hline
\multicolumn{1}{c}{\textbf{Experiment}}    & \textbf{Settings}                  \\ \hline \hline
\multicolumn{1}{|c|}{\pq{*}{*}}         & \multicolumn{1}{l|}{
\begin{tabular}[l]{l}
$t_q$ was set to 160 and 170 for the weights and activations, \\
respectively. $t_p, \Delta t_p,\text{ and } n$ were set to $140,5, \text{ and }4$, respectively.
\end{tabular}
} \\ \hline
\multicolumn{1}{|c|}{\qp{*}{*}} & \multicolumn{1}{l|}{
\begin{tabular}[l]{l}
$t_q$ was set to 140 and 150 for the weights and activations, \\
respectively. $t_p, \Delta t_p,\text{ and } n$ were set to 155,5, and 4, respectively.
\end{tabular}
} \\ \hline
\multicolumn{1}{|c|}{\q{*}} & \multicolumn{1}{l|}{
\begin{tabular}[l]{l}
$t_q$ was set to 140 and 150 for the weights and activations, respectively.
\end{tabular}
} \\ \hline
\end{tabular}
\caption{\ianUpdate{Summary of the hyperparameters used for super resolution}}
\label{tbl:super_resolution}
\end{table}


ESPCN is a fully convolutional network which is trained to upsample image patches of size (17, 17) to (48, 48).
During testing, ESPCN is used to upsample images of size (170, 170) to (480, 480). Due to the unequal sizes of training and testing images, the binary mask $\mat{M}_{\mat{x},s}$ resulting from training using unstructured
activation pruning cannot be directly applied for testing.
\ianUpdate{Accounting for this mismatch, we replicate the binary mask $\mat{M}_{\mat{x},s}$ along the height and width axis of each feature map to align with the testing size requirements.}

\ianUpdate{We calculate the {performance density}} (PD) for this task as \ianUpdate{the peak signal-to-noise ratio (PSNR) divided by the log of the combined size of the weights and activations as measure in megabits (Mb) such that the units are PSNR / log(Mb).}
We use $\log$ here as PSNR is a logarithmic domain metric.
The memory footprint of the activations is calculated assuming an input size of (1, 170, 170).

\subsection{Objection Detection on Pascal VOC} 

We use~\citep{jjfaster2rcnn} as the foundation of our object detection experiments and train Faster RCNN~\citep{o_fasterrcnn} using ResNet101~\citep{he2016deep}
on the Pascal VOC dataset~\citep{o_pascalvoc} for all experiments.
ResNet101 is pretrained on ImageNet dataset~\citep{o_imagenet}.
We follow a experiment setup similar to Appendix~\ref{apn:cifar10}, but we do not quantize or prune the first 3 residual blocks as they are frozen in the original paper~\citep{o_fasterrcnn}.
We also freeze all pre-trained batch normalization layers.
However, these limitations can be overcome with group normalization or synchronized batch normalization according to \cite{he2019rethinking}.
We set $E$ to 9, $q_u$ to 0, $q_l$ to 1, and $T$ to $32$ in each experiment. 

\begin{table}[H]
\centering
\begin{tabular}{cl}
\hline
\multicolumn{1}{c}{\textbf{Experiment}}    & \textbf{Settings}                  \\ \hline \hline
\multicolumn{1}{|c|}{\pq{*}{*}}         & \multicolumn{1}{l|}{
\begin{tabular}[l]{l}
$t_q$ was set to 7.2 and 7.5 for the weights and activations, \\
respectively. $t_p, \Delta t_p,\text{ and } n$ were set to $3, 0.5, \text{ and }4$, respectively.
\end{tabular}
} \\ \hline
\multicolumn{1}{|c|}{\qp{*}{*}} & \multicolumn{1}{l|}{
\begin{tabular}[l]{l}
$t_q$ was set to 5.2 and 5.5 for the weights and activations, \\
respectively. $t_p, \Delta t_p,\text{ and } n$ were set to 5.7, 0.5, and 4, respectively.
\end{tabular}
} \\ \hline
\multicolumn{1}{|c|}{\q{*}} & \multicolumn{1}{l|}{
\begin{tabular}[l]{l}
$t_q$ was set to 5.2 and 5.5 for the weights and activations, respectively.
\end{tabular}
} \\ \hline
\end{tabular}
\caption{\ianUpdate{Summary of the hyperparameters used for object detection}}
\label{tbl:object_detection}
\end{table}


Since the image sizes differ between each batch during training, we cannot obtain a fixed size binary mask $\mat{M}_{\mat{x},s}$ on activation of the convolution layers.
Alternatively, we apply channel-wise pruning on convolution activations. 

We calculate the {performance density} (PD) for this task as {mean average precision (mAP) divided by the combined size of the weights and activations as measure in gigabits (Gb)}, such that the units are mAP/Gb.
The memory footprint of the activations is calculated assuming an input size of (3, 480, 720).

\subsection{Pix2Pix on Facades and CycleGAN on Horse2zebra} 

We use Pix2Pix~\citep{pix2pix} and CycleGAN~\citep{o_cyclegan} \xinyu{in our} generative adversarial network (GAN) experiments.
While Pix2Pix uses a UNet architecture~\citep{unet}, CycleGAN uses a variant of ResNet with deconvolution layers.
We use an experimental setup similar to Appendix~\ref{apn:cifar10}, but we use 12-bit quantization over the activations of CycleGAN.
In addition to skipping first and last layer, we also skip both the second and second-to-last layer of CycleGAN as well as the inner-most UNet layer of Pix2Pix.
We set $q_u$ to 0 and $q_l$ to 1 in each Pix2Pix experiment and $q_u$ to 0.0001 and $q_l$ to 0.9999 in each CycleGAN experiment.
All other hyperparameters are the same between Pix2Pix and CycleGAN.
We set $E$ to $300$ and $T$ to $128$. 

\begin{table}[H]
\centering
\begin{tabular}{cl}
\hline
\multicolumn{1}{c}{\textbf{Experiment}}    & \textbf{Settings}                  \\ \hline \hline
\multicolumn{1}{|c|}{\pq{*}{*}}         & \multicolumn{1}{l|}{
\begin{tabular}[l]{l}
$t_q$ was set to 280 and 290 for the weights and activations, \\
respectively. $t_p, \Delta t_p,\text{ and } n$ were set to $100, 15, \text{ and }4$, respectively.
\end{tabular}
} \\ \hline
\multicolumn{1}{|c|}{\qp{*}{*}} & \multicolumn{1}{l|}{
\begin{tabular}[l]{l}
$t_q$ was set to 110 and 120 for the weights and activations, \\
respectively. $t_p, \Delta t_p,\text{ and } n$ were set to 130, 15, and 4, respectively.
\end{tabular}
} \\ \hline
\multicolumn{1}{|c|}{\q{*}} & \multicolumn{1}{l|}{
\begin{tabular}[l]{l}
$t_q$ was set to 110 and 120 for the weights and activations, respectively.
\end{tabular}
} \\ \hline
\end{tabular}
\caption{\ianUpdate{Summary of the hyperparameters used for object detection}}
\label{tbl:gans}
\end{table}

\begin{figure*}[]
    \centering
	\scalebox{0.5}{
		\includegraphics{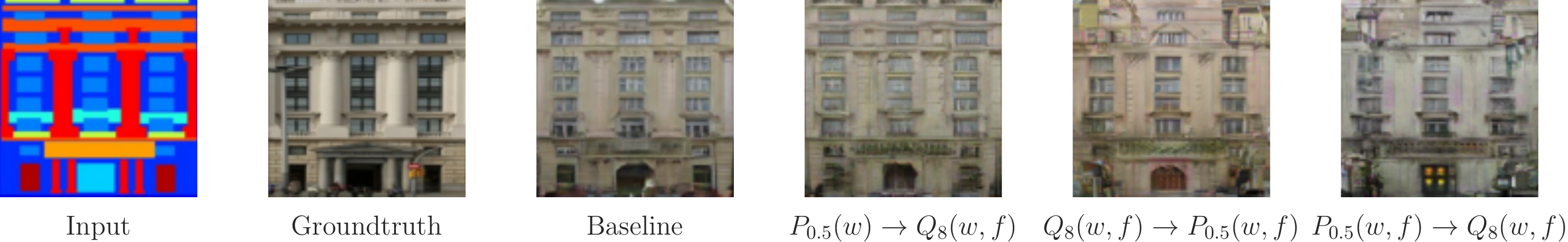}
	}
\caption{Examples of generated images from Pix2Pix.}
\label{fig:pix2pix_demos}
    \vspace{-1em}
\end{figure*}
\begin{figure*}[]
    \centering
	\scalebox{0.5}{
		\includegraphics{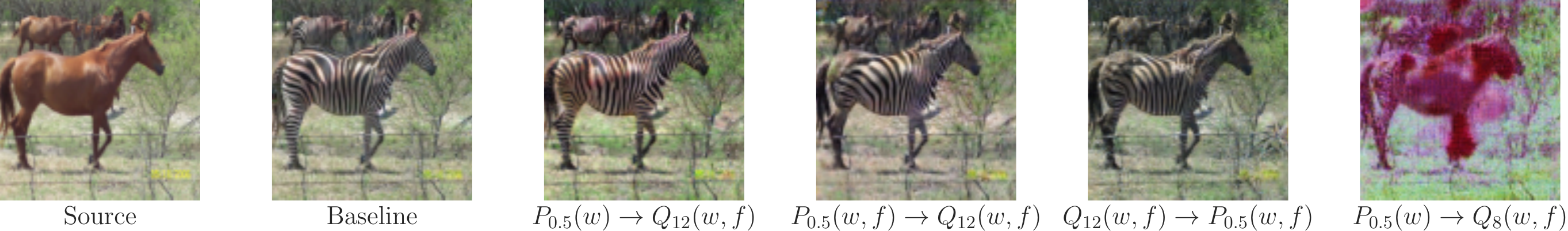}
	}
	\caption{Examples of generated images from CycleGAN.}
	\label{fig:cyclegan_demos}
	\vspace{-1em}
\end{figure*}

We calculate the performance density (PD) for each task as the Fréchet Inception Distance (FID) divided by the combined size of the weights and activations as measured in bits such that the units are 1/(FID*bits).
The memory footprint of the activations is calculated assuming an input size of (3, 256, 256).
Figures~\ref{fig:pix2pix_demos} and~\ref{fig:cyclegan_demos} respectively show examples of images generated from Pix2Pix and CycleGAN.

\end{document}